\documentclass{article}

\usepackage[preprint]{neurips_2025}

\usepackage[utf8]{inputenc} 
\usepackage[T1]{fontenc}    
\usepackage{hyperref}       
\usepackage{url}            
\usepackage{booktabs}       
\usepackage{amsfonts}       
\usepackage{nicefrac}       
\usepackage{microtype}      
\usepackage{xcolor}         
\usepackage{graphicx}
\usepackage{amsmath}
\usepackage{amssymb}
\usepackage{float} 
\usepackage{multirow}

\title{Object-Centric World Models Meet Monte Carlo
Tree Search}

\author{%
  Rodion Vakhitov \\
  MIPT\\
  Moscow, Russia \\
  \texttt{rodionvahitoff@yandex.ru} \\
  \And
  Leonid Ugadiarov \\
  AIRI \& MIPT \\
  Moscow, Russia \\
  \AND
  Aleksandr Panov \\
  AIRI \& MIPT \& FRC CSC RAS \\
  Moscow, Russia\\
}

\begin{document}

\maketitle

\begin{abstract}
In this paper, we introduce ObjectZero, a novel reinforcement
learning (RL) algorithm that leverages the power of object-level representations to
model dynamic environments more effectively. Unlike traditional approaches that
process the world as a single undifferentiated input, our method employs Graph
Neural Networks (GNNs) to capture intricate interactions among multiple objects.
These objects, which can be manipulated and interact with each other, serve as
the foundation for our model’s understanding of the environment.
We trained the algorithm in a complex setting teeming with diverse, interactive objects, demonstrating its ability to effectively learn and predict object dynamics.
Our results highlight that a structured world model operating on object-centric representations can be successfully integrated into a model-based RL algorithm utilizing Monte Carlo Tree Search as a planning module.
\end{abstract}
\section{Introduction}
Research in cognitive neuroscience and psychology shows that our minds naturally segment sensory input into discrete objects, aiding perception, generalization, and adaptability \cite{ferraro2023symmetry,hawkins2017theory}. Unlike traditional approaches that encode whole scenes \cite{hafner2019learning}, object-centric models focus on interactions between elements. Advances in unsupervised learning now allow automatic discovery and separation of objects from raw input \cite{jiang2023object,chakravarthy2023spotlight,kirilenko2024}, embedding a bias toward objectness. This leads to more interpretable and generalizable representations, aligning with cognitive principles and improving performance in downstream tasks.

\begin{figure}[t]
\includegraphics[width=\textwidth]{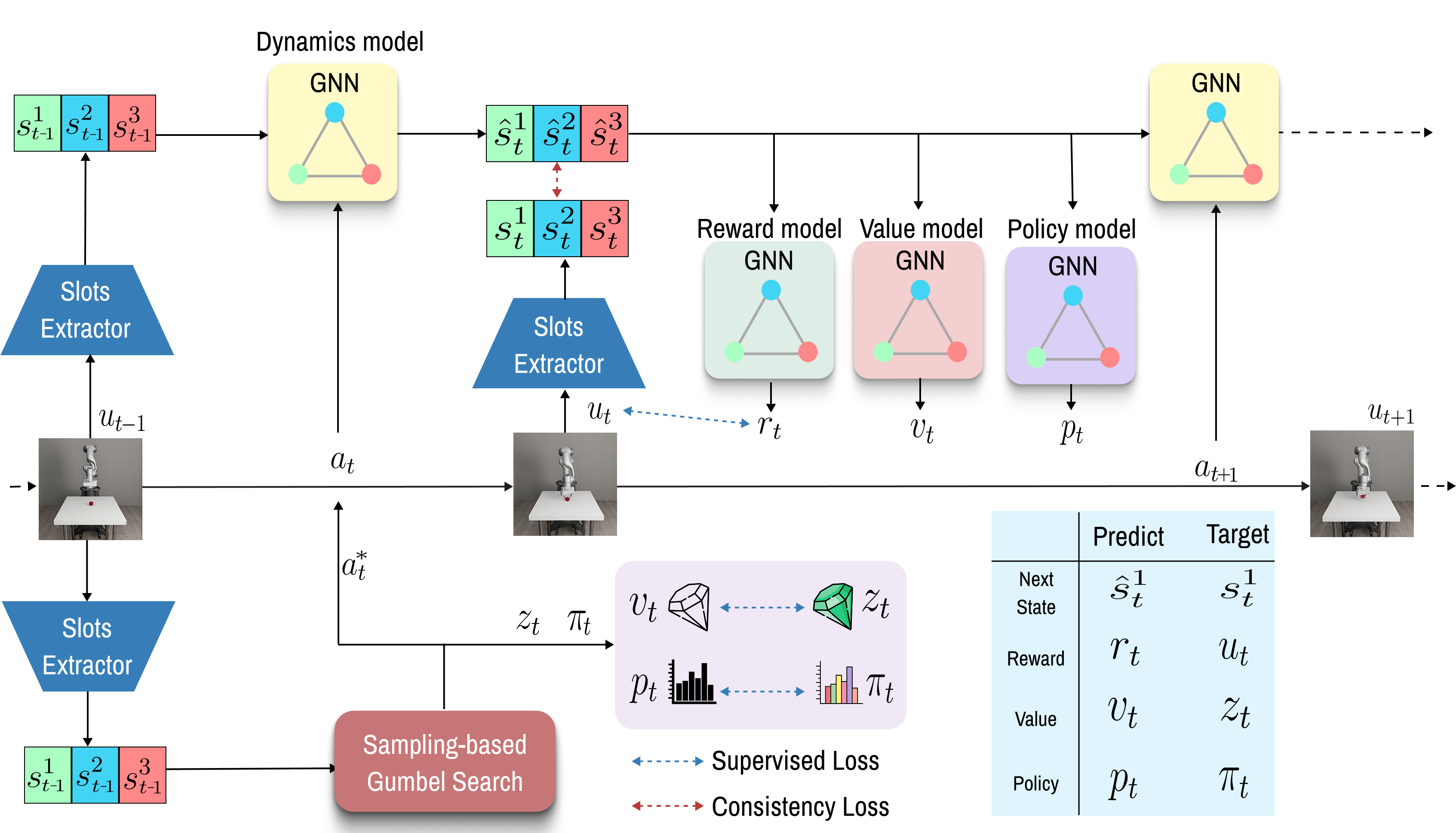}
\caption{ObjectZero training overview. The slots extractor encodes observations into object-centric representations. Separate GNN models predict next-state slots, rewards, policy and value outputs. Gumbel search generates target policy $\pi_t$ and value $z_t$.} \label{fig1}
\end{figure}

Interest in model-based reinforcement learning (MBRL) approaches \cite{DBLP:journals/corr/abs-2006-16712} has been steadily growing, driven by their ability to improve sample efficiency and generalization in reinforcement learning. The success of MBRL depends on model quality - especially in complex environments. By decomposing environments into objects and their interactions, factorized world models offer more accurate and generalizable dynamics than monolithic, entangled models. This structure supports compositionality, task transfer, and aligns with human cognitive principles.

Object-centric representations offer significant advantages in robotics and MBRL by providing structured views of the environment that enhance perception and decision-making~\cite{ugadiarov2025c}. In robotics, agents need to identify and distinguish objects and their properties to interact effectively. Such representations help robots generalize to new contexts, manipulate unfamiliar objects, and reason about action consequences \cite{devin2018deep}. By embedding such structured representations, also referred to as slots, robots gain improved autonomy and robustness, which is especially useful in complex, dynamic environments like robotic manipulation, autonomous driving, and embodied AI.

This paper introduces ObjectZero, an object-centric reinforcement learning algorithm that integrates a structured world model with online policy learning. The world model utilizes a pretrained, frozen object-centric encoder (i.e., slots extractor) based on the SLATE \cite{singh2022illiterate} or DINOSAUR \cite{seitzer2023bridging} architecture, which employ Slot Attention \cite{locatello2020object} to extract sets of object representations from high-dimensional visual inputs. The choice depends on the visual complexity of the environment: SLATE is preferred for synthetic or stylized settings, while DINOSAUR
is used for more realistic visual environments. A GNN aggregates information across these object representations to predict environment rewards, state values, dynamics and policy.

\section{Related Work}
\subsection{Object-Centric Representation Learning}
In recent years, object-centric representation learning methods have gained significant traction, aiming to extract and track individual entities from raw visual input without supervision. A central component in many of these approaches is Slot Attention~\cite{locatello2020object,frolov2025}, which employs a normalized cross-attention mechanism to iteratively bind slots to different regions in the input. Building on this foundation, SAVi \cite{kipf2022conditional} and SAVi++ \cite{elsayed2022savi++} are extensions of Slot Attention for processing video data, introducing temporal consistency by predicting optical flow and depth, respectively. SLATE and STEVE \cite{singh2022simple} achieve high-quality reconstruction through hybrid architectures.The SLATE incorporates a dVAE \cite{van2017neural} for internal feature extraction, a \textsc{GPT}-like transformer~\cite{ramesh2021zero} for decoding, and a slot-attention module to group features associated with the same object. STEVE differs from SLATE in that it operates on real-world video. DINOSAUR is trained to reconstruct not the original image based on the slots obtained using Slot Attention, but rather some encoded features of this image using the pretrained DINO ViT \cite{amir2021deep}.
\subsection{Object-Centric Representations and RL}
Modern model-based reinforcement learning methods—such as MuZero \cite{schrittwieser2020mastering}, EfficientZeroV2 \cite{wang2024efficientzero}, DreamerV3 \cite{hafner2023mastering}, TD-MPC \cite{hansen2024tdmpc} and STORM \cite{zhang2023storm} -- have significantly improved sample efficiency and planning in visual environments.
These methods achieve this by learning internal dynamics models and leveraging techniques such as Monte Carlo Tree Search (MCTS) \cite{coulom2006efficient}, Model Predictive Path Integral control~\cite{williams2015modelpredictivepathintegral}, and transformer-based architectures.
Recent work explores the integration of object-centric representations into MBRL. COBRA \cite{watters2019cobra} trains a dynamics model on the slot space learned by the MONet \cite{burgess2019monet} model, integrating intrinsically motivated exploration to optimize data acquisition and subsequent training. FOCUS \cite{ferraro2023focus} decomposes observations into separate latent object representations via an object encoder and decoder that learn to segment the scene using masking. OC-STORM \cite{zhang2025objects} predicts dynamics using a spatiotemporal transformer that handles both object and visual representations. We do not compare our method with COBRA, as it does not account for interactions between objects, which are essential for our task. FOCUS and OC-STORM are also excluded due to their reliance on manually annotated object masks, which contradicts our setting. Nevertheless, these approaches highlight the growing recognition of structured representations as a key component for scalability and interpretability of MBRL. In contrast, ROCA \cite{ugadiarov2025relationalobjectcentricactorcritic} is closely aligned with our setting: it learns an object-centric world model from unsupervised slot representations and explicitly models object interactions via graph neural networks integrated into the critic.

In parallel to these model-based approaches, recent research has also explored model-free reinforcement learning with object-centric representations. In OCRL \cite{yoon2023investigation} authors introduce a novel approach where a transformer encoder serves as a pooling mechanism within the PPO algorithm \cite{schulman2017proximal}, enabling the integration of object-level representations derived from input images by diverse object-centric models. In another study~\cite{stanic2023learning}, the authors propose the OC-CA and OC-SA algorithms, which employ Slot Attention as an object-centric feature extractor and investigate their performance and generalization abilities.

\section{Background}
\subsection{Reinforcement Learning}
Reinforcement Learning can be formulated as a Markov Decision Process (MDP) \cite{bellman1957markovian}. An MDP in this context is formalized as a tuple $(\mathcal{S}, \mathcal{A}, T, R, \gamma)$, where $s \in \mathcal{S}$ represents states, $a \in \mathcal{A}$ denotes actions, $T\colon \mathcal{S} \times \mathcal{A} \rightarrow \mathcal{S}$ is the transition function, $R\colon \mathcal{S} \times \mathcal{A} \rightarrow \mathbb{R}$ is the reward function associated with a particular task, and $\gamma$ is the discount factor. 
The goal of reinforcement learning is to find the optimal policy $\pi^* = \arg\max_{\pi} \mathbb{E}_{s_{t+1} \sim T(\cdot|s_t, a_t),\ a_{t+1} \sim \pi(\cdot|s_{t+1})} \left[ \sum_{i=0}^{\tau} \gamma^i R(s_t, a_t) \right]$ for all $s_0$ where $\tau$ is the number of time steps.

\subsection{EfficientZeroV2}
\begin{table}[h]
\caption{Comparison of component models and implementations in EZ-V2 and ObjectZero}
\label{tab:comparison-models}
\centering
\begin{tabular}{|l|l|l|}
\hline
\textbf{Component} & \textbf{Model} & \textbf{Implementation} \\
\hline
\multicolumn{3}{|c|}{\textbf{EZ-V2}} \\
\hline
Dynamics model & $\hat{s}_{t+1} \sim p_\phi(\hat{s}_{t+1} \mid s_{t}, a_{t})$ & CNN \\
Encoder & $s_t \sim p_\theta(s_t \mid o_t)$ & CNN \\
Reward model & $r_t \sim p_\psi(r_t \mid s_t)$ & CNN and MLP \\
Value model & $v_t \sim p_\xi(v_t \mid s_t)$ & CNN and MLP \\
Policy model & $p_t \sim p_\omega({p}_t \mid s_t)$ & CNN and MLP \\
\hline
\multicolumn{3}{|c|}{\textbf{ObjectZero}} \\
\hline
Dynamics model & $\hat{s}_{t+1}^i \sim p_\phi(\hat{s}_{t+1}^i \mid \hat{s}_{t}^i, a_{t})$ & GNN \\
Encoder & $\bar{s}_t = \text{Slots Extractor}(o_t)$ & SLATE (frozen) or DINOSAUR (frozen)\\
Reward model & $r_t \sim p_\psi(r_t \mid \bar{s}_t)$ & GNN and MLP \\
Value model & $v_t \sim p_\xi(v_t \mid \bar{s}_t)$ & GNN and MLP \\
Policy model & $p_t \sim p_\omega(p_t \mid \bar{s}_t)$ & GNN and MLP \\
\hline
\end{tabular}
\end{table}

Our model builds on top of the EfficientZeroV2, a model-based reinforcement learning algorithm that significantly improves upon its predecessor, EfficientZero \cite{ye2021mastering}, by extending its capabilities to continuous action spaces, enhancing planning efficiency, and introducing more accurate value estimation methods. Like EfficientZero, EZ-V2 is based on the MuZero-style framework and learns a latent model of the environment, which is used both for planning and for learning policy and value functions.
The EfficientZero family of algorithms uses MCTS as a planning module to simulate future trajectories and guide decision-making, leveraging a learned model rather than the true environment.

All the main components of EZ-V2 and ObjectZero are presented in Table 1. The current observation is represented by $o_t$, $s_t$ is the current latent state. The encoder learns a compact state representation of the input $o_t$. The dynamic function predicts the next state $\hat{s}_{t+1}$ based on the $s_t$ current state and the action taken $a_t$. The policy model outputs the current policy $p_t$, the value predictor provides the value estimation $\hat{v}_t$ at the current state and reward predictor estimates $\hat{r}_t$. All parameters of the components are trained jointly to match the target policy, value, and reward.

\begin{equation}
\mathcal{L}_{t} = \lambda_{1} \mathcal{L}_{\mathcal{R}}(u_{t}, r_{t}) + \lambda_{2} \mathcal{L}_{\mathcal{P}}(\pi_{t}, p_{t}) + \lambda_{3} \mathcal{L}_{\mathcal{V}}(z_{t}, v_{t}) + \lambda_{4} \mathcal{L}_{\mathcal{G}}(s_{t+1}, \hat{s}_{t+1})
\end{equation}

\noindent where \( u_{t} \) denotes the environment reward, \( \pi_{t} \) is the target policy from the search and \( z_{t} \) represents the target value from the search. \( \mathcal{L}_{\mathcal{R}} \), \( \mathcal{L}_{\mathcal{P}} \) and \( \mathcal{L}_{\mathcal{V}} \) all represent supervised learning losses. \( \mathcal{L}_{\mathcal{G}} \) is the temporal consistency loss, which is calculated through the negative cosine similarity. $\lambda_{1}, \lambda_{2},\lambda_{3}, \lambda_{4}$ are total loss function coefficients. \( \mathcal{L}_{t} \) is the objective function to be optimized. The agent’s interaction experience with the environment is collected into a replay buffer. The ground truth data used in the loss functions is sampled from this buffer.

Key to EZ-V2's sample efficiency are its planning and value estimation techniques. Firstly, for continuous action spaces EfficientZeroV2 generates target policies $z_{t}$ by sampling $M$ candidate actions from a mixture of the current policy $p_t$ and a prior. For discrete action spaces each action is selected as $A = \arg\max_a \left( g(a) + \text{logits}(a) \right)$, where $g_a$ is a vector of Gumbel variables and $\text{logits}(a)$ is the logit of action $a$. The candidates are ranked using Sequential Halving, which progressively discards weaker actions, i.e., actions with lower Q-values, allocating computation to the most promising ones. This enables efficient policy improvement in continuous action spaces.

To improve learning stability, EZ-V2 use Search-based Value Estimation (SVE), which computes the target value as the empirical average over simulated trajectories:

\begin{equation}
\hat{V}_S(s_t) = \frac{1}{N} \sum_{n=0}^N \left( \sum_{t=0}^{H(n)} \gamma^t \hat{r}_{t} + \gamma^{H(n)} \hat{V}(\hat{s}_{H(n)}) \right),
\end{equation}
where $N$ denotes the number of simulations, $\hat{V}_n(s_0)$ is the bootstrapped estimation of the $n$-th node expansion, $H(n)$ is a search depth of $n$-th iteration.

However, early in training or when using stale data, model error may degrade SVE accuracy. We therefore use a mixed value target, combining multi-step TD and SVE:

\begin{equation}
V_{\text{mix}} =
\begin{cases}
\sum_{i=0}^{l-1} \gamma^k u_{t+i} + \gamma^l v_{t+l} & \text{if } i_t < T_1 \text{ or } i_s > |D| - T_2  \\
\hat{V}_S(s_t) & \text{otherwise}
\end{cases}
\end{equation}
In this context, \( l \) denotes the horizon for the multi-step temporal difference (TD) method. The variable \( i_t \) represents the current training iteration, while \( T_1 \) corresponds to the initial training phase. The index \( i_s \) refers to the position of the sampled data from the replay buffer. The size of the buffer is given by \( |D| \), and \( T_2 \) defines the threshold used to evaluate the staleness of the sampled data.

\section{ObjectZero}
Figure \ref{fig1} illustrates the architectures of ObjectZero. It is composed of modular components designed around object-centric representations, enabling structured reasoning and improved generalization. Each module operates on object representations, allowing the agent to model dynamics, policy, and rewards at the level of individual entities.

\subsection{Slots Extractor}
The slots extractor takes an image-based observation $o_t$ as input and extracts object representations as an unordered set of low-dimensional vectors $\bar{s}_t = \{s^1_t, \dots, s^K_t\}$, where $K$ is a non-learned parameter which equals to the maximum possible number of objects in the image. The SLATE or DINOSAUR model can not guarantee the fixed order of object representations due to the stochastic nature of the slot-attention module. To enforce the order of object representation during the episode, we preinitialize the slots of the slots extractor model for the next step $t+1$ with the current object representations $\bar{s}_{t}$. We pretrain the SLATE and DINOSAUR on the dataset of observations collected in the environment by a uniform random policy and freeze it. Appendix ~\ref{app:slate} and Appendix ~\ref{app:dinosaur} provide information on the hyperparameters of SLATE and DINOSAUR, respectively.

\subsection{Dynamics Model}
We define the dynamics model like a graph neural network architecture, following an approach similar to that of C-SWMs \cite{Kipf2020Contrastive}. The network consists of two components: an edge model $\text{edge}_T$ and a node model $\text{node}_T$. These models take as input a factored state representation $s_t = (s^1_t, \ldots, s^K_t)$ and an action $a_t$, and predict the  next state. The MSE is used as the consistency loss.

\begin{equation}
\hat{s}^{i+1}=\texttt{node}_{T}(s_{t}^{i},a_{t}^{i},\sum_{i\neq j}\texttt{edge}_{T} (s_{t}^{i},s_{t}^{j},a_{t})).
\end{equation}

\subsection{Policy model, Reward model and Value model}

The policy, reward and value model have almost the same architecture as the dynamics model, but they do not depend on actions in their edge models or node models. We sum up object embeddings returned by the node models and feed the result into the MLP to produce the scalar value.

\begin{equation}
\begin{cases} 
\text{embed}_{V/P/R}^{i} = node_{V/P/R}\left(s_{t}^{i}, \sum_{i\neq j} edge_{V/P/R}(s_{t}^{i}, s_{t}^{j})\right) \\ 
v_{t}/p_{t}/r_{t} = \text{MLP}\left(\sum_{i=1}^{K} \text{embed}_{V/P/R}^{i}\right) 
\end{cases}
\end{equation}

The indices V, P and R refer to the value model, the policy model and the reward model, respectively. For continuous action spaces, the policy output $p_t$ is further processed to parameters of a Gaussian distribution.

\section{Environments}
\paragraph{Causal World}We conduct experiments on the Object Reaching task in the CausalWorld~\cite{ahmed2021causalworld} environment. In this task, a fixed target object and several distractor objects are placed randomly within the scene. The agent operates a tri-finger robot, where only one finger is moveable and must be used to reach the target object to earn a positive reward and complete the task. If the finger touches a distractor object first, the episode ends with zero reward. The action space is defined by the three continuous joint positions of the moveable finger. The agent does not receive any proprioceptive input, so it must learn to control the finger solely from visual observations.

\paragraph{Robosuite}In the Block Lifting task in Robosuite framework \cite{robosuite2020}, a cube is positioned on a tabletop in front of a single robotic arm. The objective for the agent is to manipulate the arm to lift the cube above a predefined height. We use a setup with a dense reward function. Our experiments utilize the Panda robot model.

\section{Experiments}

\begin{figure}[H]
\includegraphics[width=\textwidth]{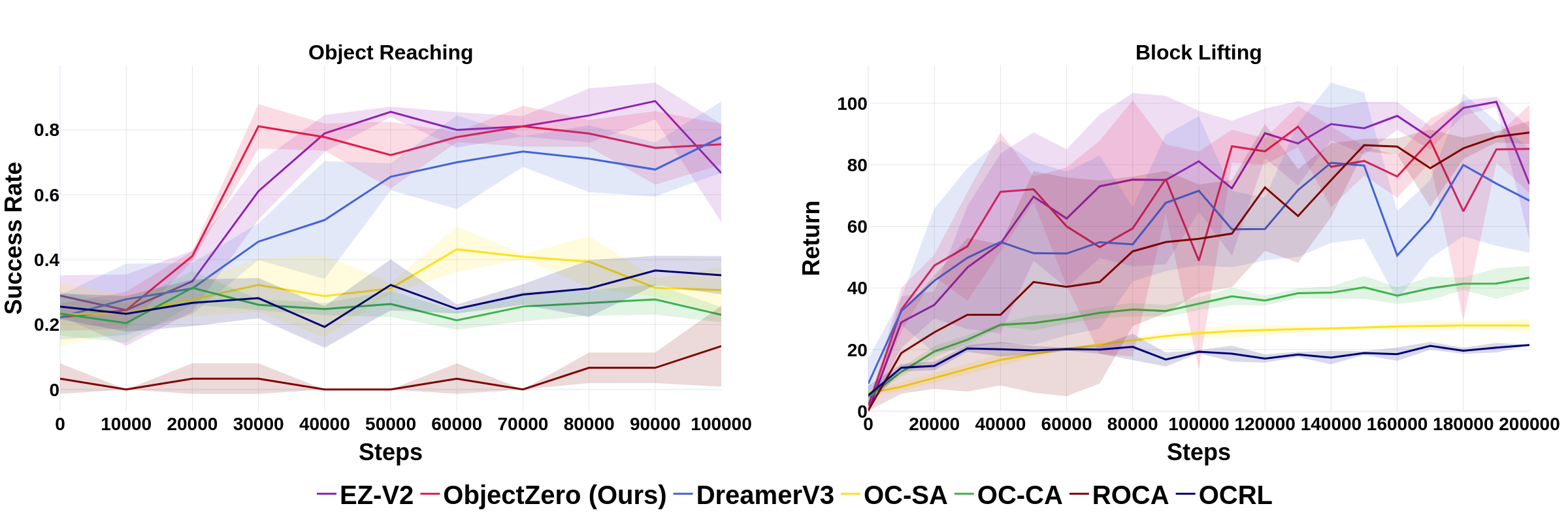}
\caption{Success rate and return averaged over 30 episodes and three seeds for ObjectZero, EZ-V2, DreamerV3, ROCA, OCRL, OC-CA and OC-SA in the Object Reaching Task in Causal World and the Block Lifting Task in Robosuite. Shaded areas indicate standard deviation} \label{fig2}
\end{figure}

\begin{figure}[H]
\includegraphics[width=\textwidth]{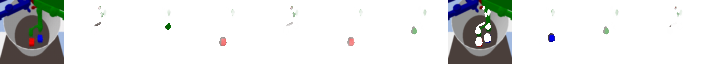}
\caption{Examples of observations and attention maps produced by the SLATE model in the Causal World Object Reaching task.} \label{fig3}
\end{figure}

\begin{figure}[H]
\includegraphics[width=\textwidth]{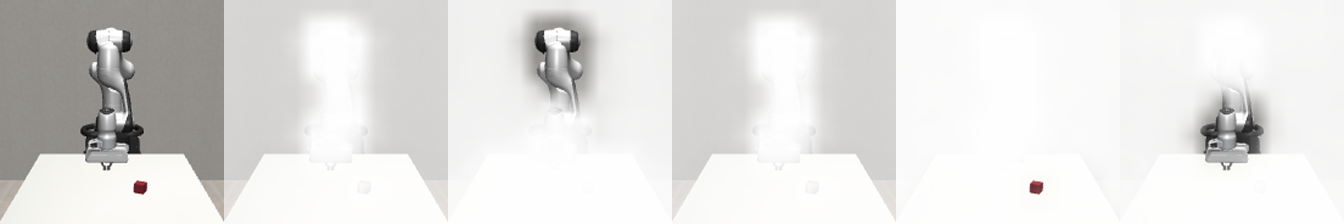}
\caption{Examples of observations and attention maps produced by the DINOSAUR model in the Robosuite Block Lifting task.} \label{fig4}
\end{figure}

We evaluate ObjectZero by comparing it to a model-free object-centric approach built on PPO, utilizing the same pre-trained and frozen SLATE and DINOSAUR model for extracting features. To aggregate the individual object representations into a format compatible with PPO's value and policy networks, we employ a Transformer encoder \cite{vaswani2017attention} as a pooling mechanism. For this setup, we adopt the Transformer-based PPO architecture from the OCRL baseline \cite{yoon2023investigation}. Additionally, we include other object-centric model-free baselines, specifically the self-attention OC-SA and cross-attention OC-CA variants proposed in \cite{stanic2023learning}. These methods train object-centric feature extractors online during policy learning.

As monolithic MBRL baselines, we include DreamerV3 \cite{hafner2023mastering} and EfficientZeroV2 \cite{wang2024efficientzero}. For the object-centric MBRL baseline, we consider ROCA \cite{ugadiarov2025relationalobjectcentricactorcritic}, which also leverages the same pre-trained and frozen SLATE and DINOSAUR model for object representation.
For the DreamerV3 baseline, we use the $\texttt{medium}$ preset for configuration parameters, with the $\texttt{train\_ratio}$ parameter increased to 128.
For OCRL, OC-SA, and OC-CA, we perform a grid search over a set of hyperparameters to improve performance.
For ROCA, we use the hyperparameters proposed by the authors for this task in the paper.
For the baseline EZ-V2, we employ the default hyperparameters specified in the original paper.
ObjectZero uses the same hyperparameters as the monolithic EZ-V2, except for $\texttt{max\_grad\_norm}$, which is decreased to 0.5. Additionally, for Block Lifting task in ObjectZero $\texttt{learning\_rate}$ is is set to 0.02. ObjectZero's implementation is directly based on the codebase of EZ-V2. For Object Reaching task in ObjectZero, ROCA and OCRL we use SLATE model (Figure \ref{fig3}), whereas in Block Lifting, we use DINOSAUR (Figure \ref{fig4}).

Figure~\ref{fig2} presents how the success rate and return depends
on the number of training steps for ObjectZero and the baseline methods on the Object Reaching  and Block Lifting task. In the Object Reaching task ObjectZero demonstrates faster convergence compared to all baselines, with the exception of EZ-V2, whose performance is comparable. In the Block Lifting task, ObjectZero outperforms OC-SA, OC-CA, and OCRL, slightly surpasses DreamerV3 and ROCA, and achieves nearly the same performance as EZ-V2. 

\section{Limitations}
Despite the advances in modern unsupervised object-centric learning, most state-of-the-art approaches still face significant challenges when it comes to decomposing complex, realistic scenes into distinct objects. We consider this limitation a major barrier to deploying our algorithm in arbitrary, real-world environments. Additionally, since we represent the environment’s state as a fully connected graph of object states, which is processed by graph neural networks, the computational complexity of ObjectZero becomes quadratic with respect to the number of slots produced by the object-centric extractor. Both of these factors can significantly slow down training in practical, object-rich scenarios—particularly in real-world robotics settings, where efficiency and scalability are crucial for learning from interaction.

\section{Conclusion}
In this work, we introduced ObjectZero, a model-based reinforcement learning algorithm that integrates object-centric representations with a structured world model built upon a pretrained SLATE and DINOSAUR encoder and Graph Neural Networks.
We demonstrated that ObjectZero effectively learns policy, outperforms existing object-centric approaches, and performs on par with the monolithic baseline in the task requiring reasoning over objects.
We show that structured dynamics and policy models operating on object-centric representations can be learned within MCTS-based MBRL algorithms.
As future work, we plan to explore the joint training of the object-centric encoder and world model components, as well as the application of ObjectZero in more realistic and high-dimensional environments where object discovery remains a major challenge.

\section{Acknowledgments}
The research was carried out using the infrastructure of the Shared Research Facilities
«High Performance Computing and Big Data» (CKP «Informatics») of FRC CSC RAS (Moscow).

\bibliographystyle{plainnat}
\bibliography{mybibliography}

\appendix

\section{SLATE} \label{app:slate}
\begin{table}[h]
\centering
\caption{Hyperparameters for the SLATE}
\label{tab:causalworld_hyperparameters}
\begin{tabular}{lll}
\toprule
\multirow{12}{*}{Learning}
 & Training dataset size & 1000000 \\
 & Temp. cooldown & 1.0 to 0.1 \\
 & Temp. cooldown steps & 30000 \\
 & LR for DVAE & 0.0003 \\
 & LR for CNN Encoder & 0.0001 \\
 & LR for Transformer Decoder & 0.0003 \\
 & LR warm-up steps & 30000 \\
 & LR half time & 250000 \\
 & Dropout & 0.1 \\
 & Clip & 0.05 \\
 & Batch size & 32 \\
 & Epochs & 100 \\
\midrule
\multirow{1}{*}{DVAE}
 & Vocabulary size & 4096 \\
\midrule
\multirow{1}{*}{CNN Encoder}\
 & Hidden size & 64 \\
\midrule
\multirow{4}{*}{Slot Attention}
 & Iterations & 3 \\
 & Slot heads & 1 \\
 & Slot dim. & 192 \\
 & MLP hidden dim. & 192 \\
\midrule
\multirow{3}{*}{Transformer Decoder}
 & Layers & 4 \\
 & Heads & 4 \\
 & Hidden dim. & 192 \\
\bottomrule
\end{tabular}
\end{table}

\section{DINOSAUR} \label{app:dinosaur}
\begin{table}[h]
\centering
\caption{Hyperparameters for DINOSAUR}
\label{tab:dinosaur_hyperparameters}
\begin{tabular}{lll}
\toprule
\multirow{13}{*}{Learning} 
 & Training dataset size & 300000 \\
 & Training steps & 500000 \\
 & Batch size & 64 \\
 & LR warm-up steps & 10000 \\
 & Peak LR & 0.0004 \\
 & Exp. decay half-life & 100000 \\
 & ViT Architecture & ViT-B \\
 & Feature dim. & 768 \\
 & Patch size & 8 \\
 & Gradient norm clipping & 1.0 \\
 & Image/Crop size & 224 \\
 & Cropping strategy & Full \\
 & Tokens & 784 \\
\midrule
\multirow{3}{*}{Decoder}
 & Type & MLP \\
 & Layers & 4 \\
 & MLP hidden dim. & 1024 \\
\midrule
\multirow{4}{*}{Slot Attention}
 & Iterations & 3 \\
 & Slots & 5 \\
 & Slot dim. & 64 \\
 & MLP hidden dim. & 512 \\
\bottomrule
\end{tabular}
\end{table}

\end{document}